\PassOptionsToPackage{table}{xcolor}
\documentclass[conference]{IEEEtran}
\IEEEoverridecommandlockouts

\usepackage[table]{xcolor}

\usepackage{cite}

\usepackage{amsmath,amssymb,amsfonts}
\usepackage{algorithmic}
\usepackage{graphicx}
\usepackage{textcomp}

\usepackage{url}
\usepackage{booktabs}
\usepackage{multirow}

\usepackage{listings}

\IEEEoverridecommandlockouts

\begin{document}

\title{Evaluating Retrieval-Augmented Large Language Models for Causal Discovery in Cerebral Palsy\\
}

\author{
\IEEEauthorblockN{
Amana Batool\textsuperscript{1},
Zhangchi Qiu\textsuperscript{1},
Christopher P. Carty\textsuperscript{2},\\
Michael Schwartz\textsuperscript{3},
Alan Wee-Chung Liew\textsuperscript{1,\(\dagger\)}
}
\IEEEauthorblockA{
\textsuperscript{1}School of ICT,
\textsuperscript{2}School of Medicine and Dentistry\\
Griffith University, Gold Coast, Australia\\
\textsuperscript{3}Department of Orthopedic Surgery,
University of Minnesota, Minneapolis, USA\\
amana.batool@griffithuni.edu.au,
\quad a.liew@griffith.edu.au
}
\thanks{\(\dagger\) Corresponding author.}
}

\maketitle

% \begin{abstract}
\begin{abstract}
Understanding how neuromuscular impairments affect gait and functional outcomes is a key goal of cerebral palsy research. Identifying these causal relationships requires careful interpretation of clinical data, domain knowledge, and modeling assumptions. Large language models (LLMs) offer a potential source of this background knowledge, but their responses may lack the specialized information and supporting evidence needed for reliable causal judgments. Retrieval from scientific literature provides supplementary domain knowledge and evidence to inform these judgments. We investigate this approach through a comparative study of retrieval-free LLMs, text-based retrieval, and graph-based retrieval, using a curated corpus of cerebral palsy literature and two causal discovery benchmarks involving metabolic power and gait kinematics. Results show that retrieval improves performance on the metabolic power task but does not outperform retrieval-free reasoning on the gait kinematics task, indicating improved causal discovery in some settings, but inconsistent benefits across tasks, and that additional context does not necessarily improve performance. These findings highlight both the value and limitations of scientific literature as a source of knowledge for LLM-based causal discovery in cerebral palsy.
\end{abstract}


\section{Introduction}

Gait difficulties in children with cerebral palsy (CP) involve relationships among neuromuscular impairments, movement patterns, and functional outcomes~\cite{clewes2024neuromuscular}. Causal models provide a framework for investigating these relationships and have been used to examine metabolic power during walking and gait outcomes after orthopedic surgery~\cite{gill2023causal,steele2022causal}. When randomized controlled trials are infeasible or unethical, observational data provide an alternative basis for investigating causal relationships~\cite{nichol2010challenging,steeger2021common}.
However, identifying causal structure from observational data remains challenging because statistical associations alone do not uniquely determine the underlying causal graph~\cite{pearl2009causal,spirtes2001causation}.
Domain knowledge is therefore important for interpreting clinical variables and informing the assumptions needed for causal discovery.

Large language models (LLMs) provide a potentially scalable source of such background knowledge. Through pretraining on large collections of textual data, LLMs acquire information about real-world entities, scientific mechanisms, and commonly reported relationships~\cite{choi2022lmpriors, long2023causal}. Previous work has shown that LLMs can use variable names and natural language descriptions to predict pairwise causal relationships and construct causal graphs~\cite{kiciman2024causal}. This capability provides a way to incorporate textual knowledge into causal discovery alongside approaches based on observational data. However, an LLM’s ability to identify familiar relationships does not ensure reliable judgments about specialized clinical variables~\cite{feng2025reliability}. Its predictions also draw on parametric knowledge whose supporting sources are difficult to inspect, making it challenging to assess whether a proposed relationship has an adequate scientific basis. These limitations motivate investigating whether relevant scientific literature can strengthen the background knowledge available to LLMs for causal discovery in CP.

\begin{figure}[t]
    \centering
    \includegraphics[width=\columnwidth]{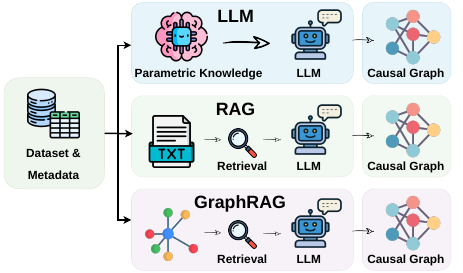}
    \caption{Overview of LLM-based causal discovery using parametric knowledge alone, text-based retrieval (RAG), or graph-based retrieval (GraphRAG).}
    \label{fig:overview}
\end{figure}

Retrieval-augmented generation (RAG) supplements an LLM’s parametric knowledge with relevant information retrieved from external sources during inference~\cite{lewis2020retrieval,ram2023context}. For causal discovery in CP, scientific publications provide supplementary knowledge about clinical variables, biomechanical mechanisms, and reported relationships. We consider two retrieval tools: text-based RAG, which retrieves passages or summaries~\cite{sarthi2024raptor}, and graph-based RAG (GraphRAG), which organizes entities and relationships into a knowledge graph to guide retrieval~\cite{guo2025lightrag,pmlr-v267-gutierrez25a}. GraphRAG’s relational structure offers a way to access connected evidence about variables and the mechanisms linking them. These retrieved connections provide context for causal judgments but do not themselves establish causation.

For causal discovery in CP, the usefulness of either retrieval tool depends on whether the supplied evidence supports judgments about the clinical variables being modeled. Relevant publications can describe associations, indirect effects, or relationships under different clinical conditions, which do not automatically support a direct causal edge in the target graph. The central question is therefore whether supplementing an LLM with this literature improves its causal predictions, and whether text-based and graph-based retrieval provide consistent benefits across different CP tasks.

In this study, we compare retrieval-free LLMs, text-based RAG, and GraphRAG for causal discovery in CP. Fig.~\ref{fig:overview} illustrates these three approaches and how they use parametric knowledge and retrieved evidence to predict causal graphs. We compile a domain-specific corpus of scientific literature and evaluate the approaches on two benchmarks, Metabolic Power and Gait Kinematics~\cite{gill2023causal,steele2022causal}.
To support a consistent comparison across retrieval approaches, we keep the corpus, downstream language model, candidate variable pairs, and metadata fixed. We examine the contribution of retrieval to causal discovery performance, its sensitivity to the amount of retrieved context, and differences across categories of clinical variables.

The main contributions of this study are as follows:
\begin{itemize}
    \item We provide a controlled comparison of parametric-only, text-based, and graph-based knowledge access for LLM-assisted causal discovery.
    \item We compile a scientific literature corpus covering CP-related impairments, gait, biomechanical mechanisms, interventions, and functional outcomes.
    \item We evaluated different causal discovery methods on two datasets, \textbf{Metabolic Power} and \textbf{Gait Kinematics}, by decomposing retrieval effects into changes in orientation accuracy and graph-level structural error.
\end{itemize}

\section{Related Work}
\begin{figure*}[t]
  \centering
  \includegraphics[width=\textwidth]{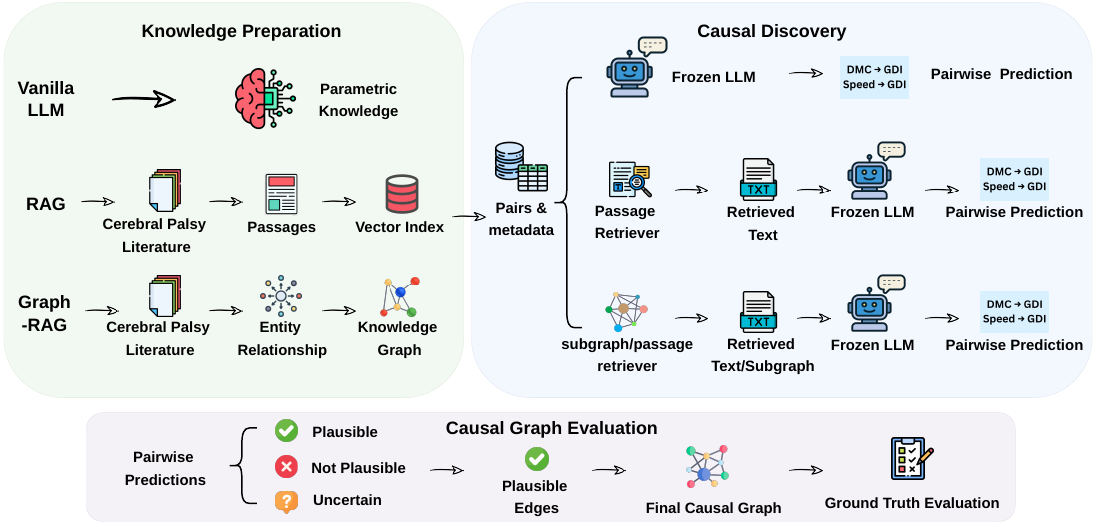}
  \caption{Comparative causal discovery workflow for vanilla LLM, RAG, and GraphRAG, showing knowledge preparation and pairwise causal prediction using a frozen LLM. Predicted plausible edges are assembled into a causal graph and evaluated against the ground truth.}
  \label{fig:framework}
\end{figure*}

\subsection{Retrieval-Augmented Generation}
Different retrieval-augmented generation (RAG) architectural designs have been proposed in recent years; for example, RAPTOR~\cite{sarthi2024raptor} retrieves hierarchical summaries, LightRAG~\cite{guo2025lightrag} combines graph and vector retrieval, and HippoRAG~\cite{pmlr-v267-gutierrez25a} uses a knowledge graph (KG) and Personalized PageRank to support associative and multi-hop retrieval. These architectures may affect causal discovery differently, as relevant evidence may recover missed causal relations, whereas incomplete, indirect, or misleading evidence may introduce false edges.

\subsection{Knowledge Retrieval-Based Causal Discovery}
Recent work has begun to connect document retrieval and structured external knowledge directly with causal discovery. LACR~\cite{zhang2024causal} integrates scientific evidence with LLM knowledge and experimental information, while IRIS~\cite{feng2025iris} combines document-extracted relations with statistical discovery. Other approaches use KG-derived priors~\cite{ban2025harnessing}, metapath-based subgraph retrieval~\cite{susanti2025paths}, or GraphRAG coupled with score-based discovery~\cite{lin2026knowledge}. CausalGraphBench~\cite{babakov2025causalgraphbench} instead benchmarks LLM-based graph construction from variable metadata. These studies primarily introduce specific knowledge-augmentation pipelines or evaluate LLMs without a controlled comparison of retrieval architectures.

\subsection{Causal Modeling in Cerebral Palsy}
Causal modeling has been applied to cerebral palsy (CP) datasets to investigate gait impairments, walking energetics, and treatment outcomes.~\cite{gill2023causal} combined a domain-informed causal graph with Bayesian additive regression trees to estimate how gait and neuromuscular impairments affect metabolic power.~\cite{steele2022causal} used a similar approach to examine motor control and postoperative gait kinematics, while~\cite{schwartz2022short} compared three causal estimators for common treatments. Our study addresses the complementary task of causal graph recovery across two CP datasets, comparing retrieval-free LLMs with text-based and graph-based retrieval.

\section{Method}

This study compares retrieval-free LLMs, RAG, and GraphRAG for causal discovery on two gait-related datasets from children with CP. Across retrieval-based methods, we keep the corpus, downstream language model, candidate variable pairs, and metadata fixed. Fig.~\ref{fig:framework} summarizes the evaluation workflow, from corpus preparation and retrieval to pairwise causal prediction and graph evaluation.

\subsection{Literature Collection}
We compile a scientific literature corpus $\mathcal{D}$ covering CP-related gait abnormalities, impairments, biomechanical mechanisms, interventions, and functional outcomes. We search \texttt{PubMed}~\cite{fiorini2017towards} and \texttt{EPMC}~\cite{ferguson2021europe} through their respective APIs using the following query, restricting the search to publications from the 15 years preceding collection:
\begin{center}
\begin{minipage}{0.82\linewidth}
\begin{lstlisting}[
    basicstyle=\ttfamily\footnotesize,
    frame=single,
    rulecolor=\color{gray},
    backgroundcolor=\color{gray!5},
    breaklines=true,
    columns=fullflexible,
    keepspaces=true,
    showstringspaces=false,
    aboveskip=0pt,
    belowskip=0pt,
    framesep=5pt
]
("cerebral palsy"[MeSH Terms]
OR "cerebral palsy"[Title/Abstract])
AND
(
    gait[MeSH Terms]
    OR "gait analysis"[Title/Abstract]
    OR "walking"[Title/Abstract]
    OR "atypical gait"[Title/Abstract]
    OR "gait impairments"[Title/Abstract]
    OR "crouch gait"[Title/Abstract]
    OR "CP gait"[Title/Abstract]
)
\end{lstlisting}
\end{minipage}
\end{center}
We downloaded article content in BioC JSON format from \texttt{PubMed} and JATS XML format from \texttt{EPMC}, then normalized it into structured JSON records containing article metadata, abstracts, and available main-text sections. References, acknowledgments, and supplementary sections are excluded from the processed full text. An LLM-based filter screens the documents using their titles, abstracts, and full text where available, assigning each document to an \textit{Accept}, \textit{Reject}, or \textit{Uncertain} category. We retain accepted documents to form the machine-readable corpus $\mathcal{D}$ for retrieval and exclude the rejected or uncertain documents. The resulting corpus contained more than 2500 documents, which form the machine-readable document corpus $\mathcal{D}$ for further use in knowledge retrieval architectures. We explicitly excluded the two documents containing ground-truth information on the causal graphs for both datasets from the corpus.

\subsection{RAG-based Pipelines}
\subsubsection{Text-Based RAG}
Retrieval-augmented generation (RAG) conditions a language model's
response on external evidence retrieved for an input query.
Given a query $q$ and a document corpus $\mathcal{D}$, a retriever $\mathcal{R}$
selects the top-$k$ passages according to a relevance function $s$:
\begin{equation}
\mathcal{R}_k(q;\mathcal{D})
=
\operatorname{TopK}_{d\in\mathcal{D}} s(q,d).
\end{equation}
The selected passages are assembled into a context 
$C_q=\operatorname{Concat}(\mathcal{R}_k(q;\mathcal{D}))$,
which is provided to the language model alongside the query.
The model generates a response according to the conditional distribution
\begin{equation}
p_{\mathrm{RAG}}(y\mid q,\mathcal{D})
=
p_{\theta}(y\mid q,C_q),
\end{equation}
where $y$ denotes the generated response and $\theta$ denotes
the model's parameters. This formulation combines the model's
parametric knowledge with query-relevant external information.
\subsubsection{GraphRAG}

Graph-based retrieval-augmented generation (GraphRAG) organizes external knowledge as a graph $\mathcal{G}=(\mathcal{V},\mathcal{E})$, where the nodes in $\mathcal{V}$ represent entities or information units and the edges in $\mathcal{E}$ encode relationships between them. Given a query $q$, a graph-aware retriever selects relevant evidence:
\begin{equation}
\mathcal{S}_q = \operatorname{Retrieve}(q,\mathcal{G}),
\end{equation}
where $\mathcal{S}_q$ contain entities, relations, paths,
subgraphs, or graph-derived summaries. The retrieved evidence,
optionally augmented with associated source passages, is converted
into a textual context $C_q=\operatorname{Serialize}(\mathcal{S}_q)$.
A reasoning language model then generates a response conditioned
on the query and this context:
\begin{equation}
p_{\mathrm{GraphRAG}}(y\mid q,\mathcal{G})
=
p_{\theta}(y\mid q,C_q),
\end{equation}
where $y$ denotes the generated response and $\theta$ denotes
the model's parameters. GraphRAG uses relational structure
to connect relevant information across the knowledge source
and support reasoning over multiple evidence units.

\begin{table*}[t!]
\caption{Causal discovery performance across the Metabolic Power and Gait Kinematics datasets.
Prec.\ and Rec.\ denote precision and recall.
Darker and lighter blue cells indicate the best and second-best distinct
values, respectively; ties are highlighted equally.}
\label{tab:cp-gait-main-results}
\centering
\small
\renewcommand{\arraystretch}{1.05}
\resizebox{\linewidth}{!}{%
\begin{tabular}{@{}llcccccccccccc@{}}
\toprule
\multirow{2}{*}{\textbf{Category}}
& \multirow{2}{*}{\textbf{Method}}
& \multicolumn{6}{c}{\textbf{Metabolic Power}}
& \multicolumn{6}{c}{\textbf{Gait Kinematics}} \\
\cmidrule(lr){3-8}\cmidrule(lr){9-14}
& & \textbf{Prec.$\uparrow$} & \textbf{Rec.$\uparrow$}
& \textbf{F1$\uparrow$} & \textbf{SHD$\downarrow$}
& \textbf{NHD$\downarrow$} & \textbf{Ratio$\downarrow$}
& \textbf{Prec.$\uparrow$} & \textbf{Rec.$\uparrow$}
& \textbf{F1$\uparrow$} & \textbf{SHD$\downarrow$}
& \textbf{NHD$\downarrow$} & \textbf{Ratio$\downarrow$} \\
\midrule
\multirow{2}{*}{Traditional} & PC
& 0.266 & 0.133 & 0.177 & 32 & 0.291 & 1.000
& 0.333 & 0.188 & 0.240 & \cellcolor{blue!10}13 & \cellcolor{blue!10}0.232 & \cellcolor{blue!10}1.000 \\
 & DAG-GNN
& 0.416 & 0.167 & 0.238 & 29 & 0.264 & 0.906
& 0.375 & 0.188 & 0.250 & 15 & 0.268 & 1.154 \\
\midrule
\multirow{8}{*}{LLM-only} & GPT-4o
& 0.500 & 0.400 & 0.444 & 30 & 0.273 & 0.938
& 0.272 & 0.188 & 0.222 & 19 & 0.339 & 1.462 \\
 & \quad + metadata
& 0.500 & 0.600 & 0.545 & 30 & 0.273 & 0.938
& 0.235 & 0.250 & 0.242 & 21 & 0.375 & 1.615 \\
%  & Qwen3-8B
% & 0.513 & 0.633 & 0.567 & 29 & 0.264 & 0.906
% & \cellcolor{blue!20}0.714 & 0.313 & 0.434 & \cellcolor{blue!10}13 & \cellcolor{blue!10}0.232 & \cellcolor{blue!10}1.000 \\
%  & \quad + metadata
% & 0.529 & 0.600 & 0.562 & 28 & 0.255 & 0.875
% & 0.454 & 0.313 & 0.370 & 17 & 0.304 & 1.308 \\
 & Qwen3-14B
& \cellcolor{blue!20}0.823 & 0.467 & 0.595 & 19 & 0.173 & 0.594
& 0.500 & 0.063 & 0.111 & 16 & 0.286 & 1.231 \\
 & \quad + metadata
& 0.636 & 0.700 & 0.666 & 21 & 0.191 & 0.656
& 0.428 & 0.188 & 0.260 & 17 & 0.304 & 1.308 \\
 & Deepseek-v4-flash
& 0.611 & 0.367 & 0.458 & 26 & 0.236 & 0.813
& 0.464 & \cellcolor{blue!10}0.813 & 0.590 & 18 & 0.321 & 1.385 \\
 & \quad + metadata
& 0.531 & 0.567 & 0.548 & 28 & 0.255 & 0.875
& \cellcolor{blue!20}0.576 & \cellcolor{blue!20}0.938 & \cellcolor{blue!20}0.714 & \cellcolor{blue!20}12 & \cellcolor{blue!20}0.214 & \cellcolor{blue!20}0.923 \\
\midrule
\multirow{2}{*}{RAG} & Naive RAG
& 0.727 & 0.533 & 0.615 & 18 & 0.164 & 0.563
& 0.470 & 0.500 & 0.484 & 16 & 0.286 & 1.231 \\
 & RAPTOR
& 0.725 & \cellcolor{blue!20}0.967 & \cellcolor{blue!20}0.828 & \cellcolor{blue!20}12 & \cellcolor{blue!20}0.109 & \cellcolor{blue!20}0.375
& 0.500 & \cellcolor{blue!10}0.813 & 0.619 & 15 & 0.268 & 1.154 \\
\midrule
\multirow{2}{*}{GraphRAG} & HippoRAG 2
& \cellcolor{blue!10}0.757 & 0.833 & \cellcolor{blue!10}0.793 & \cellcolor{blue!10}13 & \cellcolor{blue!10}0.118 & \cellcolor{blue!10}0.406
&\cellcolor{blue!10} 0.550 & 0.688 & 0.611 & \cellcolor{blue!20}12 & \cellcolor{blue!20}0.214 & \cellcolor{blue!20}0.923 \\
 & LightRAG
& 0.682 & \cellcolor{blue!10}0.933 & 0.788 & 15 & 0.136 & 0.469
& 0.520 & \cellcolor{blue!10}0.813 & \cellcolor{blue!10}0.634 & 14 & 0.250 & 1.077 \\
\bottomrule
\end{tabular}
}
\end{table*}

\subsection{Causal Graph Recovery}
Given a variable set $\mathcal{X}=\{X_1,\ldots,X_n\}$ and corresponding
metadata $\mathcal{M}=\{m_1,\ldots,m_n\}$, a causal query $q_{ij}$ is constructed
for each unordered variable pair:
\begin{equation}
q_{ij}=\operatorname{Query}(X_i,m_i,X_j,m_j),
\qquad 1\leq i<j\leq n.
\end{equation}
Each query asks whether a direct causal relationship exists between the variables and, if so, in which direction. The reasoning model predicts a label $\hat{r}_{ij}\in\{\rightarrow,\leftarrow,\varnothing\}$ from the query and an optional evidence context $C_{ij}$:
\begin{equation}
\hat{r}_{ij}
=
\operatorname{LLM}_{\theta}(P,q_{ij},C_{ij}),
\end{equation}
where $P$ specifies the causal reasoning task and $\varnothing$
denotes either no direct causal relationship or an uncertain response, which are discarded before aggregation into the final graph. The context is empty for
retrieval-free prediction and contains retrieved evidence for
retrieval-augmented prediction.The pairwise predicted plausible edges are aggregated
into a directed graph $\hat{\mathcal{G}}=(\mathcal{X},\hat{\mathcal{E}})$,
with
\begin{equation}
\begin{aligned}
\hat{\mathcal{E}}
={}&
\{(X_i,X_j):i<j,\ \hat{r}_{ij}=\rightarrow\}\\
&\cup
\{(X_j,X_i):i<j,\ \hat{r}_{ij}=\leftarrow\}.
\end{aligned}
\end{equation}
The resulting graph represents the causal relationships predicted across the variable pairs.

\subsection{Graph Evaluation}
The recovered graph
$\hat{\mathcal{G}}=(\mathcal{X},\hat{\mathcal{E}})$ is evaluated
against the ground-truth graph
$\mathcal{G}^{*}=(\mathcal{X},\mathcal{E}^{*})$
over the same variable set. 
We treat each ordered pair of distinct variables as a candidate edge and evaluate edge predictions using precision, recall, and F1. We also measure structural discrepancy using structural Hamming distance (SHD), defined here as the minimum number of edge additions, deletions, or reversals required to transform $\hat{\mathcal{G}}$ into $\mathcal{G}^{*}$, with each reversal counted as one operation.
\section{Results}
We evaluate the contribution of scientific literature retrieval to causal discovery in CP through three research questions: \textbf{RQ1}: Does literature retrieval improve causal discovery in CP compared with retrieval-free LLMs? \textbf{RQ2}: How does the amount of retrieved context affect causal discovery performance? \textbf{RQ3}: How does performance vary across categories of clinical variables?

\subsection{Experimental Setup}
\subsubsection{Datasets}
\begin{figure}[t]
    \centering
    \includegraphics[width=0.98\columnwidth]{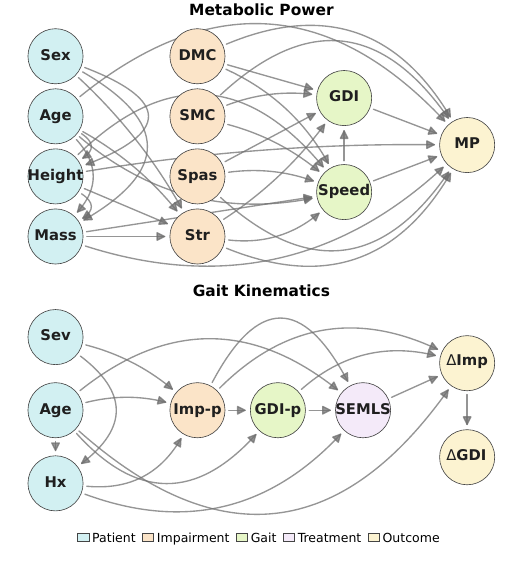}
    \caption{Ground-truth causal graphs for the Metabolic Power and Gait Kinematics datasets.}
    \label{fig:ground_truth_graphs}
\end{figure}

We evaluate the approaches on two cerebral palsy datasets, Metabolic Power~\cite{gill2023causal} and Gait Kinematics~\cite{steele2022causal}, using the expert-designed ground-truth causal graphs provided in Fig.~\ref{fig:ground_truth_graphs}. Metabolic Power represents relationships among patient characteristics, neuromuscular impairments, gait, and metabolic power during walking. Gait Kinematics represents relationships among baseline impairments, gait, orthopedic treatment, and changes in clinical outcomes. Together, these benchmarks allow us to evaluate causal discovery for two distinct tasks in CP research. The expert-designed graphs were assessed by testing their implied conditional independencies against observational data~\cite{steele2022causal,gill2023causal}.
\subsubsection{Metrics}
We evaluate graph recovery using standard metrics from prior causal discovery work~\cite{du2025causal} including Precision, Recall, F1 score, Structural Hamming Distance (SHD), Normalized Hamming Distance (NHD), and Ratio. NHD is the normalized version of SHD. Lower NHD and SHD indicate better structural recovery while accounting for graph size. Ratio is NHD relative to PC on the same dataset; lower values indicate higher structural accuracy.

\subsubsection{Baselines}
We compare three groups of methods:
\textbf{(i) Data-driven:} PC~\cite{spirtes2001causation} and DAG-GNN~\cite{yu2019dag};
\textbf{(ii) LLM-only:} GPT-4o~\cite{hurst2024gpt}, Qwen3-14B~\cite{yang2025qwen3}, and DeepSeek-v4-flash~\cite{xu2026deepseek};
\textbf{(iii) Retrieval-augmented:} Naive RAG~\cite{laitenberger2025stronger}, RAPTOR~\cite{sarthi2024raptor}, HippoRAG 2~\cite{pmlr-v267-gutierrez25a}, and LightRAG~\cite{guo2025lightrag}.
We selected the LLM baselines to represent diverse model families, parameter scales, and reasoning capabilities, while fixing DeepSeek-V4-Flash as the downstream reasoner to isolate the contribution of different retrieval architectures.

\subsection{\textbf{RQ1: Does literature retrieval improve causal discovery in CP compared with retrieval-free LLMs?}} 
Table~\ref{tab:cp-gait-main-results} shows that retrieval contributes differently across the two CP tasks. On Metabolic Power, all retrieval-based methods achieve higher F1 scores and lower SHD than the retrieval-free Deepseek model with metadata. RAPTOR achieves the highest F1 score of 0.828 and the lowest SHD of 12, compared with 0.548 and 28, respectively, for that retrieval-free baseline. Within the graph-based approaches, LightRAG achieves higher recall than HippoRAG 2 but lower precision and higher SHD. Thus, identifying more ground-truth relationships does not necessarily produce a graph with fewer errors. On Gait Kinematics, the retrieval-free Deepseek model with metadata achieves the highest F1 score of 0.714. LightRAG achieves the highest F1 score among retrieval-based methods at 0.634, while HippoRAG 2 matches the retrieval-free model's lowest SHD of 12. These results show that retrieval does not improve on the metadata-informed retrieval-free model for this task, although the relative performance of the retrieval approaches depends on the metric. Across the two benchmarks, literature retrieval improves causal discovery for Metabolic Power but provides no consistent advantage for Gait Kinematics. The results therefore support evaluating retrieved evidence within the clinical task being modeled, without assuming that either text-based RAG or GraphRAG is consistently preferable.

\subsection{\textbf{RQ2: How does the amount of retrieved context affect causal discovery performance?}}

\begin{figure}[t]
    \centering
    \includegraphics[width=\columnwidth]{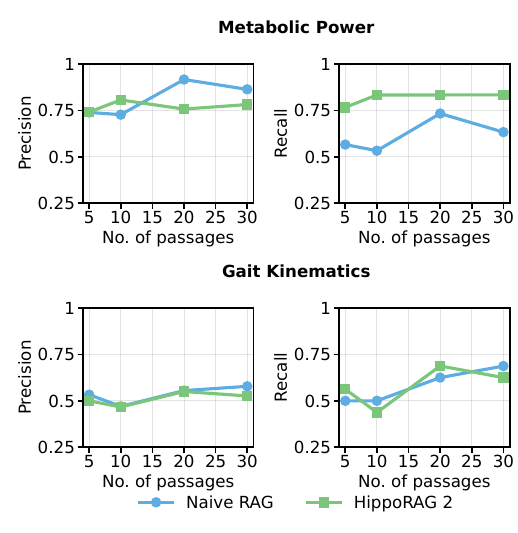}
    \caption{Effect of retrieval depth on causal graph recovery
    for Metabolic Power and Gait Kinematics datasets.}
    \label{fig:retrieval-depth-ablation}
\end{figure}

We vary the number of retrieved passages across 5, 10, 20, and 30 for Naive RAG and HippoRAG 2. 
The results revealed a non-monotonic relationship between context size and causal-graph recovery, with the preferred retrieval budget varying across methods and datasets. 
Fig.~\ref{fig:retrieval-depth-ablation} shows that increasing context does not consistently improve precision or recall. On Metabolic Power, HippoRAG 2 achieves higher recall than Naive RAG at every evaluated passage count. Its recall increases from 5 to 10 passages and remains unchanged at 20 and 30. Naive RAG achieves its highest precision and recall at 20 passages, followed by a decline at 30. At 20 and 30 passages, Naive RAG has higher precision than HippoRAG 2 but lower recall. 

On Gait Kinematics, the relative performance changes with the passage count. HippoRAG 2 has higher recall at 5 and 20 passages, while Naive RAG has higher recall at 10 and 30. HippoRAG 2's recall peaks at 20 passages and declines at 30, whereas Naive RAG peaks at 30 passages. These patterns show that the effect of additional literature context varies across both the retrieval method and the CP task. One possible explanation for the declines at larger passage counts is that additional passages introduce irrelevant or redundant information, increasing noise in the context and making it harder for the model to identify evidence relevant to the causal query. Providing more passages therefore does not guarantee more accurate causal predictions.

\subsection{\textbf{RQ3: How does performance vary across categories of clinical variables?}}
\begin{figure}[t]
    \centering
    \includegraphics[width=\columnwidth]{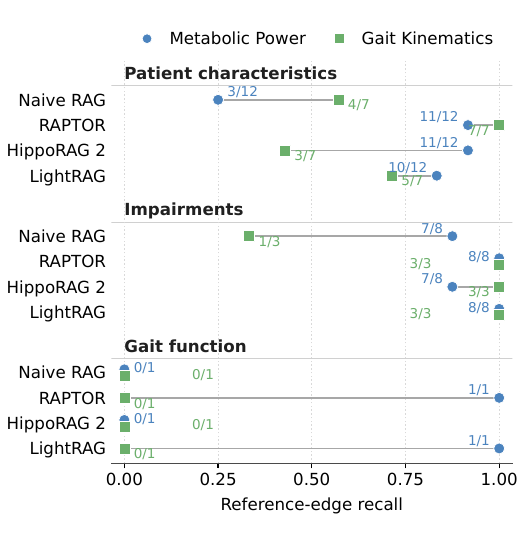}
    \caption{Reference-edge recovery recall for the Metabolic Power and Gait Kinematics datasets across different categories. Each point represents the number of recovered true edges compared to the actual true edges.}
    \label{fig:category_recovery}
\end{figure}

We examine three clinical variable categories shared by the benchmarks: Patient Characteristics, Impairments, and Gait Function. Fig.~\ref{fig:category_recovery} reports the proportion of ground-truth directed edges correctly identified in each category. This analysis shows which categories of relationships the retrieval-based methods identify within each CP task.
For Patient Characteristics, RAPTOR and HippoRAG 2 each correctly identify 11 of 12 ground-truth edges on Metabolic Power. On Gait Kinematics, RAPTOR identifies all seven edges, while HippoRAG 2 identifies three. LightRAG identifies 10 of 12 and five of seven edges, respectively, exceeding Naive RAG on both benchmarks. For Impairments, RAPTOR and LightRAG identify all ground-truth edges in both benchmarks. HippoRAG 2 identifies seven of eight edges on Metabolic Power and all three on Gait Kinematics, while Naive RAG identifies seven of eight and one of three, respectively.

The Gait Function category contains one ground-truth edge in each benchmark. RAPTOR and LightRAG correctly identify this edge on Metabolic Power, whereas neither Naive RAG nor HippoRAG 2 does. None of the evaluated retrieval methods identify the edge on Gait Kinematics. This finding is specific to the two evaluated edges and does not establish a general limitation across gait-related relationships. Overall, RAPTOR achieves the highest or joint-highest recall in each evaluated category, but even this approach misses a relationship in Gait Kinematics. The category results illustrate that performance within a CP task varies across the clinical relationships being predicted.

\section{Conclusion}
This work presented a controlled comparison of parametric, text-based, and graph-based retrieval methods for LLM-assisted causal discovery.  The results show that retrieval augmentation should not be treated as inherently superior to closed-book LLM reasoning. Instead, its benefit depends on the dataset and architecture.
Overall, our findings indicate that hierarchical and graph-structured retrieval can offer advantages over flat passage retrieval, but these advantages depend on whether the retrieved context adds causal information that complements information available in the data and in the LLM's parametric knowledge. This study shows that access to scientific literature alone does not guarantee accurate causal graph recovery in cerebral palsy. Retrieval improves recovery on the Metabolic Power benchmark but does not outperform metadata-informed closed-book reasoning on Gait Kinematics, while additional passages do not consistently improve performance. The dataset-dependent gains and non-monotonic effects of passage count show that retrieval alone is insufficient, but do not explain the sources of recovery errors. Future work should examine whether these errors arise from variable interpretation, insufficient or misleading retrieved evidence, or confusion between direct and mediated relationships. Evaluation across additional reader models and domains would establish how broadly the observed patterns hold.

\section*{Acknowledgments}
This research was supported by the MRFF-funded Motion Connect project (MRFFRD11000028). The first author acknowledges financial support from the Griffith University Postgraduate Research Scholarship and the Griffith University International Postgraduate Research Scholarship.
\section*{Ethical Considerations}
This study uses previously collected, de-identified
clinical data, performs secondary analysis, and provides references to the original studies. During manuscript preparation, LLM-based assistance was used for grammar correction, writing refinement, and clarity improvement. They were not used to generate datasets, results, or final scientific conclusions. All research concepts, ideas, and analyses were developed and conducted by the authors.

\bibliographystyle{IEEEtran}
\bibliography{custom}

\end{document}